\documentclass[11pt,a4paper]{article}

\usepackage[utf8]{inputenc}
\usepackage[T1]{fontenc}
\usepackage[english]{babel}
\usepackage{amsmath,amsfonts,amssymb,amsthm}
\usepackage{graphicx}
\usepackage[caption=false]{subfig}
\usepackage{booktabs}
\usepackage{array}
\usepackage{url}
\usepackage{hyperref}
\usepackage[margin=2.5cm]{geometry}
\usepackage{algorithm}
\usepackage{algorithmic}
\usepackage{lineno}
\usepackage{xcolor}
\usepackage{enumerate}

\usepackage{times}
\usepackage{mathptmx}

\usepackage{setspace}
\usepackage{fancyhdr}
\theoremstyle{definition}

\usepackage{caption}
\hypersetup{
    colorlinks=true,
    linkcolor=blue,
    filecolor=magenta,      
    urlcolor=cyan,
    citecolor=red,
    pdftitle={Novel Approaches to Artificial Intelligence Development Based on the Nearest Neighbor Method},
    pdfauthor={I.I. Priezzhev, D.A. Danko, A.V. Shubin},
    pdfsubject={Machine Learning, Artificial Intelligence},
    pdfkeywords={nearest neighbor, hierarchical clustering, self-organizing maps, machine learning}
}

\begin{document}

\begin{titlepage}
\centering

\vspace*{2cm}
{\LARGE\textbf{Novel Approaches to Artificial Intelligence Development Based on the Nearest Neighbor Method}}

\vspace{2cm}

{\large
I.I. Priezzhev$^{1,2}$, D.A. Danko$^{1,2}$, A.V. Shubin$^{1,2}$
}

\vspace{1cm}

{\normalsize
$^1$National University of Oil and Gas «Gubkin University», Moscow, Russia\\
$^2$IPLab LLC, Moscow, Russia
}

\vspace{2cm}

{\large\textbf{Abstract}}

\vspace{0.5cm}

\begin{minipage}{0.8\textwidth}
\small
Modern neural network technologies, including large language models, have achieved remarkable success in various applied artificial intelligence applications, however, they face a range of fundamental limitations. Among them are hallucination effects, high computational complexity of training and inference, costly fine-tuning, and catastrophic forgetting issues. These limitations significantly hinder the use of neural networks in critical areas such as medicine, industrial process management, and scientific research. This article proposes an alternative approach based on the nearest neighbors method with hierarchical clustering structures. Employing the k-nearest neighbors algorithm significantly reduces or completely eliminates hallucination effects while simplifying model expansion and fine-tuning without the need for retraining the entire network. To overcome the high computational load of the k-nearest neighbors method, the paper proposes using tree-like data structures based on Kohonen self-organizing maps, thereby greatly accelerating nearest neighbor searches. Tests conducted on handwritten digit recognition and simple subtitle translation tasks confirmed the effectiveness of the proposed approach. With only a slight reduction in accuracy, the nearest neighbor search time was reduced hundreds of times compared to exhaustive search methods. The proposed method features transparency and interpretability, closely aligns with human cognitive mechanisms, and demonstrates potential for extensive use in tasks requiring high reliability and explainable results.
\end{minipage}

\vspace{1cm}

{\small\textbf{Keywords:} approximate nearest neighbor, hierarchical clustering, self-organizing maps, machine learning, geophysical applications}

\vfill

{\small
Corresponding author: I.I. Priezzhev (email: priezzhev.i@ivanplab.ru)
}

\end{titlepage}

\section{Introduction}

Modern artificial intelligence systems, particularly large language models (LLMs) based on neural networks, demonstrate impressive capabilities. For example, large language models such as GPT, PaLM, and LLaMA can generate coherent text, perform translations, answer questions, and solve complex problems \cite{brown2020language}. However, these approaches exhibit significant conceptual limitations, including hallucination effects, training complexity, expensive fine-tuning, and others. Overcoming these limitations is challenging without reconsidering fundamental concepts in the design of artificial intelligence systems.

In this paper, we analyze the root causes of the primary shortcomings of classical neural network technologies and propose an alternative approach for advancing artificial intelligence algorithms and architectures.

The foundational mechanism of the new approach is the k-nearest neighbors algorithm (k-NN) \cite{fix1951discriminatory}, based on directly comparing the test object with all elements of the training set. This method relies on the principle of analogy, ensuring transparency in decision-making logic and interpretation of results. Among the key advantages of the k-NN algorithm are the absence of a training phase, elimination of hallucinations, and straightforward expansion of the training dataset by adding new objects \cite{cover1967nearest}. Nevertheless, this mechanism has significant disadvantages: substantial computational load during exhaustive searches through the training set and the necessity of storing the entire dataset, which can be very large \cite{bentley1975multidimensional}.

In order to accelerate the nearest neighbors search, it is proposed to preprocess the training set by forming hierarchical tree structures, where nodes store generalized representations of similar information elements. An accelerated nearest neighbors search method based on hierarchical networks was previously proposed and tested by us for predictive tasks in seismic data interpretation \cite{priezzhev2024hierarchical}.

The main objective of this work is an in-depth investigation of this approach and its adaptation for solving a broader range of artificial intelligence tasks.

\section{Limitations of Neural Network Technologies}

Neural network technologies face several significant limitations hindering their further development:

\begin{itemize}
\item \textbf{Hallucination Effect}. This term describes situations where the model produces plausible yet false or nonsensical statements. This effect stems from statistical token prediction without direct reference to training examples. In the absence of closely related contextual or formal data during training, the model interpolates or extrapolates information, resulting in inaccurate or entirely fictitious assertions \cite{ji2023survey}. This phenomenon affects both large LLMs and smaller generative networks, limiting their application in scenarios requiring guaranteed accuracy.

    Even minor diagnostic errors in critical fields like healthcare can have severe consequences for patients, and mistrust in automated neural network forecasts complicates technology adoption in industrial process management systems \cite{vaswani2017attention}. Increasing occurrences of "false alarms" heighten the risk of rejecting AI-based decisions, reducing the overall efficiency and reliability of digital systems.

\item \textbf{Training Complexity}. Training modern models involves massive datasets and optimization of hundreds of millions or billions of parameters. According to scaling laws, the volume of data used and computational resource demands grow polynomially with parameter count and text size \cite{kaplan2020scaling}. The carbon footprint from a single training cycle of large models can reach tens of tons of CO$_2$-equivalent, raising concerns about the environmental sustainability of neural network technology development \cite{strubell2020energy}.

    Another significant issue is the high computational load and memory requirements during inference, the stage where a trained model is applied for predictions. Deploying mega-models in real-time requires server architectures with numerous GPUs (Graphics Processing Unit) or TPUs (Tensor Processing Unit) and substantial video memory, restricting neural network usage in mobile and embedded systems without employing compression or knowledge distillation \cite{zhang2016understanding}.

\item \textbf{Expensive Fine-Tuning}. Methods for parameter-efficient adaptation, such as Low-Rank Adapters (LoRA) and adapter layers, are actively used for adapting to new tasks and fine-tuning. These methods enable updating a small fraction of weights and reducing resource consumption. However, dependency on the original dataset size and model complexity remains \cite{hu2022lora}. Fully eliminating resource-intensive retraining remains unresolved.

\item \textbf{Catastrophic Forgetting}. This phenomenon occurs when neural networks rapidly lose previously acquired knowledge upon fine-tuning with new data, limiting continuous learning capabilities in dynamically changing environments. Methods involving regularization and limited-resource memory have been proposed, but a universal solution remains elusive \cite{kirkpatrick2017overcoming}.

\item \textbf{Neural Network Fragility}. Neural networks are vulnerable to adversarial attacks—small, often imperceptible perturbations to input data can lead to significant prediction errors. This vulnerability challenges the applicability of neural networks in security and autonomous control systems \cite{szegedy2013intriguing}.

\item \textbf{Model Calibration}. Modern neural networks often exhibit excessive confidence in their predictions, complicating the evaluation of uncertainty and decision-making risk \cite{guo2017calibration}. Poor calibration can result in underestimating errors and potential failures in systems sensitive to inaccurate predictions.

\item \textbf{Ethics and Fairness}. Neural network decisions heavily depend on training data quality. Dataset bias, for instance, leads to discrimination based on social or gender characteristics, as demonstrated in studies on facial recognition and text classification systems \cite{buolamwini2018gender}. Combating such artifacts necessitates rigorous data cleansing and the implementation of bias-detection algorithms.

\item \textbf{Black-Box Problem}. The opaque internal mechanisms of neural networks represent one of the main barriers to their widespread adoption. Lack of transparency and interpretability stimulates the development of Explainable AI (XAI), which aims to create methods and tools ensuring transparency and decision interpretability in complex models \cite{gunning2017explainable}. Despite significant progress, existing XAI methods are often either overly simplistic for complex architectures or reduce model accuracy, highlighting the need for further research balancing interpretability and effectiveness.

    Thus, despite remarkable achievements, neural network technologies face fundamental constraints ranging from technical and environmental to ethical and methodological challenges. Addressing these issues requires an interdisciplinary approach, integrating research efforts in algorithms, hardware development, and regulatory frameworks.
\end{itemize}

\section{Nearest Neighbors Method}

The k-NN method is a classical, non-parametric machine learning algorithm categorized as a "lazy" method because it does not require an explicit training phase \cite{cover1967nearest}. It operates by calculating distances from each new sample to all points in the training set, after which predictions are made based on labels (in classification tasks) or response values (in regression tasks) of the k closest neighbors.

This method exhibits high stability and offers several advantages:

\begin{itemize}
\item \textbf{Absence of a traditional training phase}. The k-NN method does not involve the adjustment or optimization of internal model weights. All points are immediately stored in a database, and the "intelligence" of the method resides in dynamically finding the nearest neighbors for each prediction. New examples do not require retraining; adding them directly to the dataset allows the algorithm to instantly incorporate the updated data.

\item \textbf{Outlier/anomaly detection}. By setting a threshold for the distance to the nearest neighbor, the method can identify samples dissimilar to any example in the dataset. This approach is widely used for novelty and anomaly detection tasks \cite{chandola2009anomaly}.

\item \textbf{Intuitive explainability}. Predictions are directly modeled through explicitly provided analogies, allowing users to provide examples of neighbors to justify the algorithm's choices. This facilitates interpretation and trust in results \cite{hastie2009elements}.

\item \textbf{Metric flexibility}. Depending on the nature of features, metrics such as Euclidean, Manhattan, Cosine, Minkowski, and others can be used, adapting the algorithm to specific data.

\item \textbf{Asymptotic optimality}. As the size of the training set approaches infinity and with an appropriate choice of k, the error of the k-NN classifier approaches the Bayes error. For k=1, the upper error bound does not exceed twice the Bayes error.
\end{itemize}

Despite its clear advantages, the k-NN method has several significant drawbacks:

\begin{itemize}
\item \textbf{Computational complexity}. A naive implementation requires $O(n \cdot d)$ operations per query (where $n$ is the number of samples, and $d$ is the dimensionality of the features), becoming a computational bottleneck for large datasets and high dimensionalities.

\item \textbf{Requirement to store the entire dataset}. All training samples must be readily accessible in memory or storage, limiting scalability and processing speed.

\item \textbf{Dimensionality sensitivity}. The "curse of dimensionality" arises when distances in high-dimensional spaces become non-informative, reducing the quality of neighbor searches. Preprocessing techniques such as feature selection or dimensionality reduction are required to mitigate this issue \cite{hastie2009elements}.

\item \textbf{Necessity of choosing the parameter k and metric}. Although k-NN lacks weight training, it has critical hyperparameters—specifically, the number of neighbors ($k$) and the distance measurement method. If $k$ is too small, the algorithm becomes excessively sensitive to noisy examples (analogous to overfitting). If $k$ is too large, predictions become overly averaged, losing structural details (analogous to excessive smoothing). Optimal selection of k and the most suitable metric typically involves cross-validation \cite{malkov2018efficient}.
\end{itemize}

\section{Hierarchical Structures of Generalized Data}

When comparing a neural network for classification tasks to a simple exhaustive search of the training dataset for the nearest neighbor, it is notable that both approaches may yield the same result. However, neural network solutions operate significantly faster since the training set is effectively "embedded" into the network as a set of weighted coefficients. In other words, a neural network represents a parallel mechanism enabling accelerated searches analogous to exhaustive searches. The primary trade-off for this speed is the loss of direct traceability of the results back to specific training examples: the answer becomes "generalized" and does not indicate individual data points.

To speed up nearest neighbor searches, various methods have been developed, such as the Hierarchical Navigable Small World (HNSW) algorithm \cite{malkov2018efficient}. The essence of this method lies in constructing multiple levels of "small worlds"—randomized graphs, each facilitating greedy navigation to the nearest nodes down to the base layer. This allows achieving sublinear, practically logarithmic complexity in high-dimensional tasks. We employ a similar approach based on constructing hierarchical decision trees \cite{priezzhev2024hierarchical}. The idea involves structurally generalizing the training set by building a multi-level search network, grouping objects according to their similarity. Such approaches are widely used in practical tasks. For instance, searching for a book in a large library does not require browsing the entire collection; it is sufficient to refer to the catalog, locating the desired book within the corresponding section or subsection—a classic example of a hierarchical search structure.

The implementation of the proposed approach begins with the first level of generalization, where clustering training objects divides the dataset into several nodes (clusters) based on similarity measures. Each cluster is subsequently subdivided into child nodes until a predetermined tree depth or a minimum cluster size is reached. Nodes containing one or a small number of objects are treated as leaves of the search tree.

If objects with different labels or significantly varying prediction targets are grouped within a single leaf, this indicates insufficient feature description or an improperly selected similarity measure. Conversely, objects with identical labels or similar predictive values within a single leaf can be replaced by a single generalized object, substantially reducing the volume of the training set.

The process of constructing a hierarchical search tree can be viewed as a training phase, while the resulting structure acts as an expanding, tree-like neural network. Input to this network is a vector describing a test object. At each hierarchy level, the nearest node is chosen, and the search continues only through its child nodes until reaching a leaf. The final exhaustive search within the leaf identifies the nearest neighbor (or several neighbors) from the training set.

Each node in this network is analogous to a "neuron" in classical architectures, with the node's level determining the degree of generalization. In terms of a library catalog, the first level corresponds to "halls" for general topics (scientific literature, fiction, etc.), the next level to shelves with books on specific disciplines, and subsequent levels to subsections refining the topic.

\begin{figure}[ht]
\centering
\includegraphics[width=0.9\textwidth]{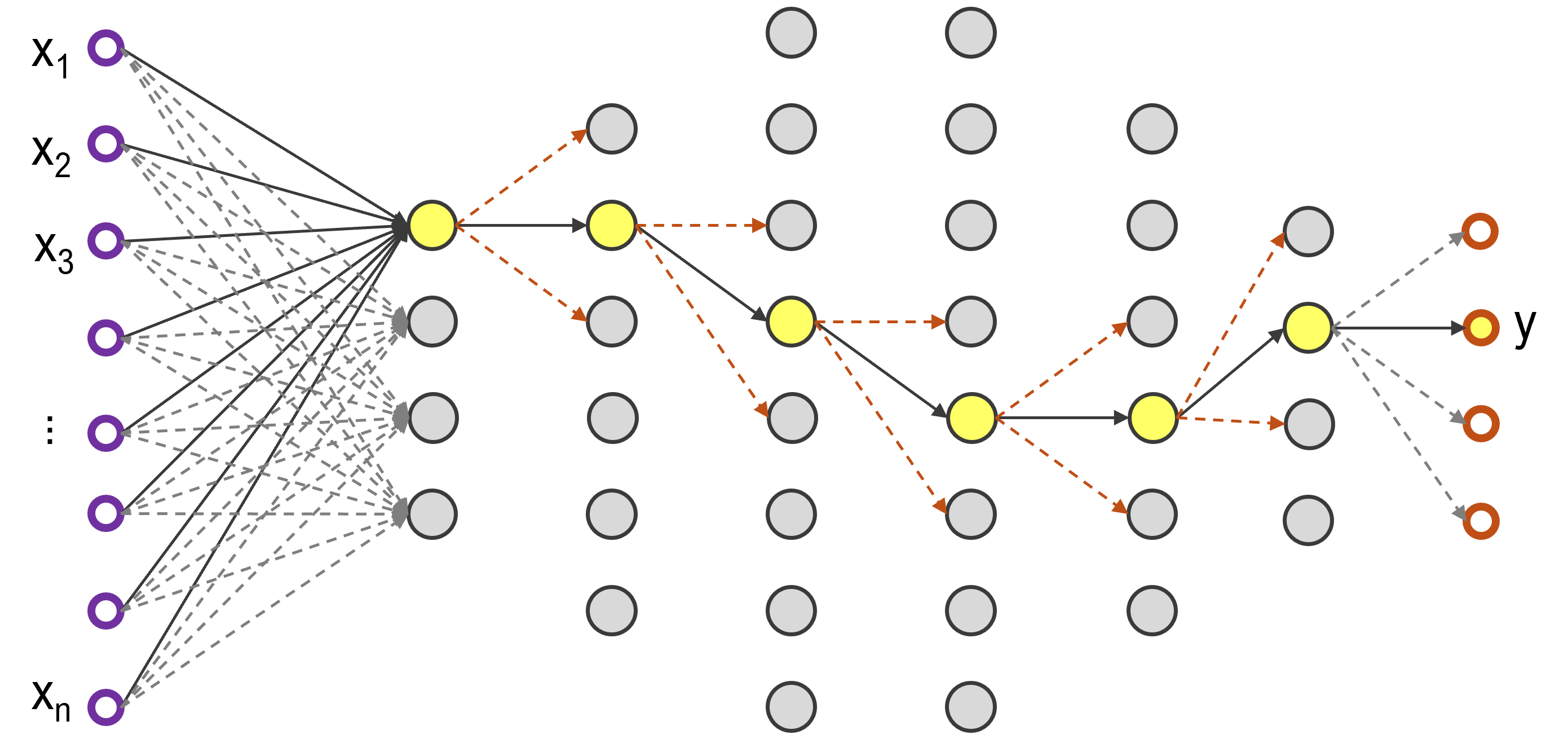}
\caption{Scheme of a hierarchical neural network for generalized data.}
\label{fig:hierarchical_scheme}
\end{figure}

The search path in a hierarchical neural network forms a chain of nodes closely matching the test object (Figure \ref{fig:hierarchical_scheme}). This fundamentally differentiates the mechanism from a standard neural network, where the input signal propagates through all neurons.

\section{Clustering the Training Set}

In constructing each level of the hierarchical search tree, we employ the Self-Organizing Maps (SOM) algorithm \cite{kohonen1995self}, using variations such as SOM 1D/2D/3D projections \cite{priezzhev2019seismic}.

Mathematically, this procedure can be described as follows:

\subsection{Initialization of SOM}

For simplicity, consider the SOM 1D clustering scenario. Let the training set contain $M$ vectors:
\begin{equation}
S_i = (s_{1i}, s_{2i}, \ldots, s_{ni}) \in \mathbb{R}^n, \quad i = 1, \ldots, M
\end{equation}

The SOM 1D algorithm involves creating a neural network of a predefined size $k$, where each node $j = 1, \ldots, k$ is associated with a weight vector of the same size:
\begin{equation}
W_j = (w_{1j}, w_{2j}, \ldots, w_{nj}) \in \mathbb{R}^n
\end{equation}

Initialization $W_j(0)$ is performed randomly, such as by selecting $k$ distinct objects from the training set.

\subsection{Iterative Learning}

At each iteration $t$, a random vector $S_i$ is selected. Then, a loop over all training set objects $S_i$ identifies the closest "winning" neuron $W_j$ using various distance metrics, typically the $L_2$ norm:
\begin{equation}
j^* = \arg\min_j \|S_i - W_j(t)\|_2
\end{equation}

Subsequently, the "winning" neuron values are updated according to:
\begin{equation}
W_j(t+1) = W_j(t) + \alpha(t)h_{\sigma(t)}(d(j,j^*))[S_i - W_j(t)]
\end{equation}

Where:
\begin{itemize}
\item $\alpha(t)$ is the learning rate, decreasing over iterations
\item $d(j,j^*) = |j - j^*|$ is the node index distance along the 1D map
\item $h_{\sigma(t)}$ is the neighborhood function, commonly Gaussian:
\end{itemize}

\begin{equation}
h_\sigma(d) = \exp\left(-\frac{d^2}{2\sigma^2}\right)
\end{equation}

or the "Mexican hat" function:
\begin{equation}
h_\sigma(d) = \left(1 - \frac{d^2}{\sigma^2}\right)\exp\left(-\frac{d^2}{2\sigma^2}\right)
\end{equation}

which adjusts nearby neighbors by a decreasing function based on distance \cite{kohonen1995self}.

\subsection{Extension to SOM 2D/3D}

For SOM 2D and SOM 3D clustering, nodes are arranged into a $k \times k$ map or $k \times k \times k$ cube. Node indices become vectors $(j_1, j_2)$ or $(j_1, j_2, j_3)$ calculated as Euclidean or Manhattan distances on the grid. Weight updates follow the same rule, accounting for multi-dimensional neighborhoods along all axes, adjusting neurons close to the "winning" neuron across all indices.

\subsection{Dimension Reduction and Prototyping}

Post-training, each input vector $S_i$ can be uniquely assigned to a neuron $j^*$. Resulting indices serve as new variables, making SOM 1D/2D/3D clustering suitable not only for grouping similar objects but also for reducing feature space dimensions. Thus, projections from n-dimensional feature spaces are mapped to new spaces of size $k$ (1D), $k \times k$ (2D), or $k \times k \times k$ (3D).

Transitioning to higher dimensions (2D, 3D) better preserves data topology and reduces quantization errors. If needed, this method can be generalized to $m$-dimensional maps, although computational costs and neighborhood function selection complexity increase with dimensionality.

\section{Algorithm Testing}

\subsection{Handwritten Digit Recognition}

To evaluate the performance and efficiency of the proposed method, we conducted tests using the MNIST (Modified National Institute of Standards and Technology) dataset. The MNIST dataset consists of 60,000 images of handwritten digits, each sized $28 \times 28$ pixels in the training set, and 10,000 images in the test set. The Euclidean metric ($L_2$ norm) was used as the similarity measure based on pixel values.

The baseline brute-force method computed distances from each of the 10,000 test samples to all 60,000 training samples, requiring more than 80 minutes on standard single-thread execution. The misclassification rate in this scenario was 3.69\% (369 errors).

\begin{figure}[ht]
\centering
\includegraphics[width=0.5\textwidth]{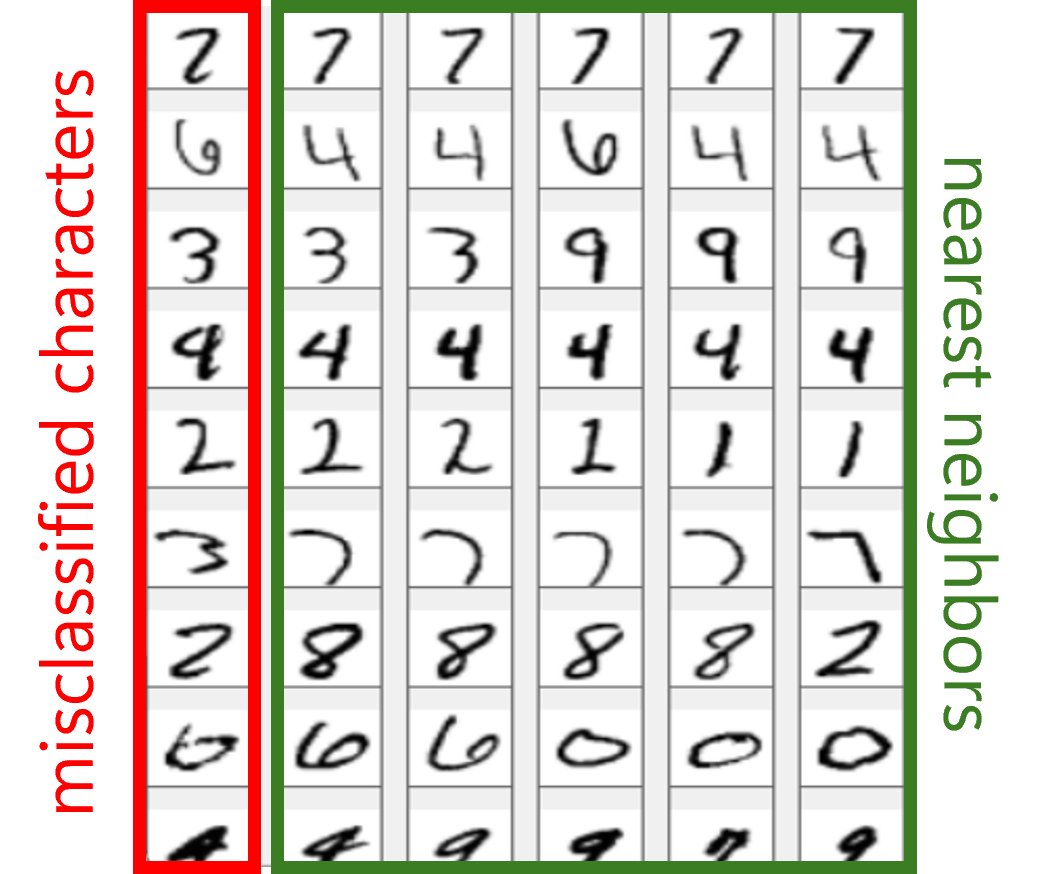}
\caption{Examples of images from the MNIST test set (10,000 samples) misclassified by the hierarchical neural network, and their nearest neighbors from the training set (60,000 samples) found using the constructed search tree.}
\label{fig:mnist_errors}
\end{figure}

Using the hierarchical method, we constructed a search tree with a branching factor of 10 at each level and a depth of 5. This approach reduced the processing time for the entire test set to approximately 0.1 minutes ($\approx$6 seconds), an acceleration of over 800 times. However, accuracy decreased to 5.64\% (564 errors) due to inaccuracies in assigning some samples to clusters. Figure \ref{fig:mnist_errors} shows examples of misclassifications along with their closest neighbors from the training set. Clearly, using more sophisticated similarity measures (e.g., based on convolutional network features) could enhance recognition accuracy.

\subsection{Simple Machine Translation}

As a second experiment, we applied our method to Russian-English subtitle translation using the publicly available OpenSubtitles v2018 (OPUS) dataset. Russian and English subtitle texts were used to create a training dataset of sentence pairs. Each sentence was transformed into a 200-length vector: first, a token dictionary was created and sorted by token frequency. Then each word was assigned a vector index within this dictionary. Sentences with fewer than 200 tokens were padded with -1, while sentences exceeding 200 tokens were truncated to the first 200.

\begin{figure}[ht]
\centering
\includegraphics[width=1\textwidth]{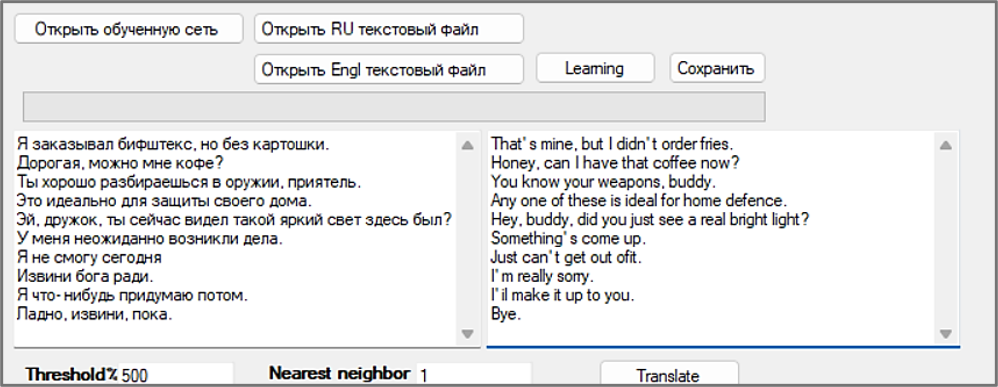}
\caption{Results of testing the hierarchical neural network as a simple subtitle translator: original sentences are shown in the right window, and corresponding translations in the left window. The similarity threshold sets the minimum similarity value for neighbor selection to avoid hallucinations, and the number of nearest neighbors determines the number of alternative translation options.}
\label{fig:translation_results}
\end{figure}

We built a hierarchical tree from these vectors using the same parameters (branching factor = 10, depth = 5). In the user interface, a sentence was entered in the source language, and the "translation" output was the closest vector from the target language measured by the $L_2$ metric (Figure \ref{fig:translation_results}). Although this approach does not match the quality of modern seq2seq models, it demonstrated the fundamental feasibility of hierarchical trees in text data processing tasks.

A key advantage of this method is its simplicity in "fine-tuning": when new sentence pairs arrive, only the token dictionary needs updating, and new vectors are processed through the training procedure (adjusting node and leaf weights) without retraining the entire structure.

Thus, hierarchical nearest-neighbor neural networks can effectively serve as a lightweight embedding method for texts, reducing the feature space while preserving local semantic relationships between words and sentences.

\section{Limitations of Hierarchical Neural Networks}

Testing of the proposed method revealed several fundamental limitations and drawbacks affecting the accuracy and performance of these models.

One primary limitation is the ambiguity in clustering objects located at cluster boundaries. In hierarchical structures, objects in boundary zones can ambiguously belong to two or more clusters, increasing recognition errors compared to exhaustive search methods. This limitation arises from the insufficient discriminative power of traditional hard clustering algorithms, such as agglomerative and divisive methods \cite{hastie2009elements, kaufman1990finding}. To mitigate this issue, soft clustering algorithms are recommended, allowing objects to simultaneously belong to multiple clusters. However, this increases computational load, with the extent of increase directly depending on the threshold of boundary object membership, which can be set manually or automatically based on data statistics.

Another significant limitation is the requirement to store the entire training dataset. Even if only object references are stored, potential duplication of boundary elements increases memory demands. This issue is particularly relevant when working with large datasets, such as texts for large language models \cite{devlin2018bert, brown2020language}. Although modern computational devices like workstations and laptops typically have sufficient storage capacity, extremely large datasets necessitate cloud storage solutions, incurring additional infrastructure costs.

Furthermore, deep hierarchical structures with numerous nodes at each level significantly increase nearest-neighbor search times. Although this process can be easily parallelized at the node level, overall computational complexity remains high, especially with considerable depth and breadth of the hierarchy \cite{beyer1999nearest}.

Hierarchical neural networks also suffer from the "curse of dimensionality." As the dimensionality of the feature space increases, distances between objects become less informative, reducing clustering efficiency and nearest-neighbor search effectiveness. This limitation is particularly evident in high-dimensional feature spaces, where hierarchical structures may not always outperform flat graph-based methods such as k-NN with diversified connections \cite{aggarwal2001surprising}. Improved results can be achieved by combining hierarchical structures with graph diversification techniques \cite{luan2020complete}.

Finally, the rigidity of hierarchical tree decisions reduces model robustness to noise and outliers, particularly in low-density data regions. This limitation complicates result interpretation and decreases overall classification stability. Contemporary approaches, such as interpretable k-NN algorithms, partially address this issue by analyzing neighbors in embedding latent spaces, enhancing robustness and improving model interpretability \cite{papernot2018deep}.

\section{Advantages of Hierarchical Neural Networks}

Despite these limitations, hierarchical neural networks exhibit several significant advantages due to their structural organization and learning principles. One primary advantage is the linear scalability of training time relative to dataset size and network geometry complexity, determined by node count per level and tree depth \cite{koller2009probabilistic}. This allows hierarchical neural networks to be effectively applied even to datasets containing billions of objects without requiring expensive supercomputing resources \cite{aggarwal2015data}.

A notable benefit is \textbf{elimination of hallucination effects} typical of generative neural models, such as Generative Adversarial Networks (GANs). With hierarchical neural networks, it is straightforward to identify when queried data is not present in the training set, as the nearest neighbor identified will exhibit a similarity parameter below a predefined threshold \cite{goodfellow2016deep}.

\textbf{High robustness} is another key feature of hierarchical neural networks. Even if a single element (e.g., neuron or coefficient within a neuron) is damaged, the network continues functioning, albeit with partial information loss in some memory branches. This significantly enhances system reliability for critical applications \cite{haykin2009neural}.

Another advantage is the capability for compatibility \textbf{analysis among training pairs}. Leaves in hierarchical structures typically group similar objects with identical or similar labels. Differences in labeling among similar objects signal insufficient vector representation, facilitating the detection of anomalous or erroneous objects \cite{jain1988algorithms}.

Additionally, hierarchical networks enable generalization of identical or similar objects, substantially reducing the size of the training dataset. Generalization is achieved by substituting multiple similar objects with a single representative object, optimizing data processing and accelerating network training \cite{bishop2006pattern}.

A crucial strength of hierarchical neural networks is their capacity for \textbf{effective fine-tuning} without catastrophic forgetting, a common issue in other neural architectures. New objects can be easily integrated into the existing tree, quickly determining their optimal hierarchical placement \cite{parisi2019continual}.

When a test object significantly differs from its nearest neighbor beyond a defined threshold, it forms a new leaf node in the hierarchy, treating the object as a new type distinct from known classes. Additionally, analyzing property trends within a leaf can theoretically predict the existence of new, undiscovered objects or classes \cite{han2011data}.

Hierarchical neural networks can also \textbf{learn effectively from minimal data}, even starting from a single object. Sequentially adding objects and adaptively forming new clusters at various hierarchy levels provides high flexibility and adaptability. Clustering similarity thresholds can be automatically determined through statistical analysis of variance in object proximity across tree levels \cite{xu2005survey}.

\section{Potential Applications of Hierarchical Neural Networks}

Hierarchical neural networks have been successfully employed across various domains, demonstrating effectiveness in solving prediction and classification tasks. One relevant and successfully implemented application area is predicting reservoir properties of rocks in oil and gas fields using seismic exploration data and well log measurements \cite{priezzhev2024hierarchical}.

\begin{figure}[ht]
\centering
\includegraphics[width=\textwidth]{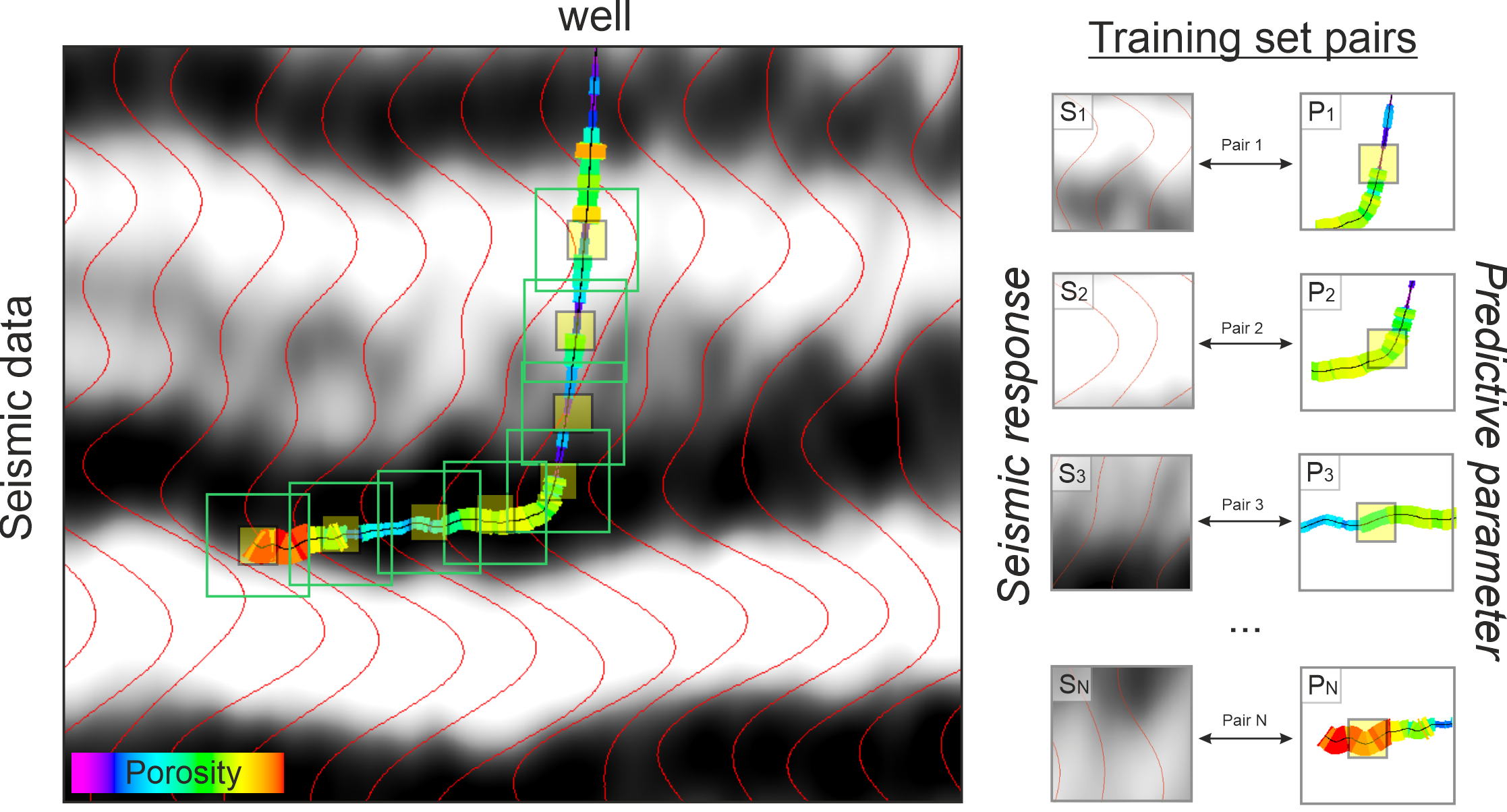}
\caption{Principle of forming a training dataset using well and seismic data.}
\label{fig:training_dataset}
\end{figure}

Training datasets in such tasks are formed by initially grouping input seismic data with target predictive parameters derived from interpreting geophysical well survey data. Training data consist of pairs such as "seismic response—predictive parameter." Frequently, continuous rock property distributions such as porosity, mineral component volume, or hydrocarbon saturation serve as the predictive parameters. Specifically, the average predictive parameter value is calculated along the well trajectory and matched to a particular cell of a regular seismic grid (Figure \ref{fig:training_dataset}).

The seismic response represents a segment (window) of the seismic wavefield with dimensions $m \times n$, where $m$ is the number of vertical samples (in time or depth scales), and $n$ is the number of adjacent traces. The center of this window corresponds to the cell location with the predictive parameter. Users set parameters $m$ and $n$, and their optimal values are empirically determined during model calibration.

\begin{figure}[ht]
\centering
\includegraphics[width=\textwidth]{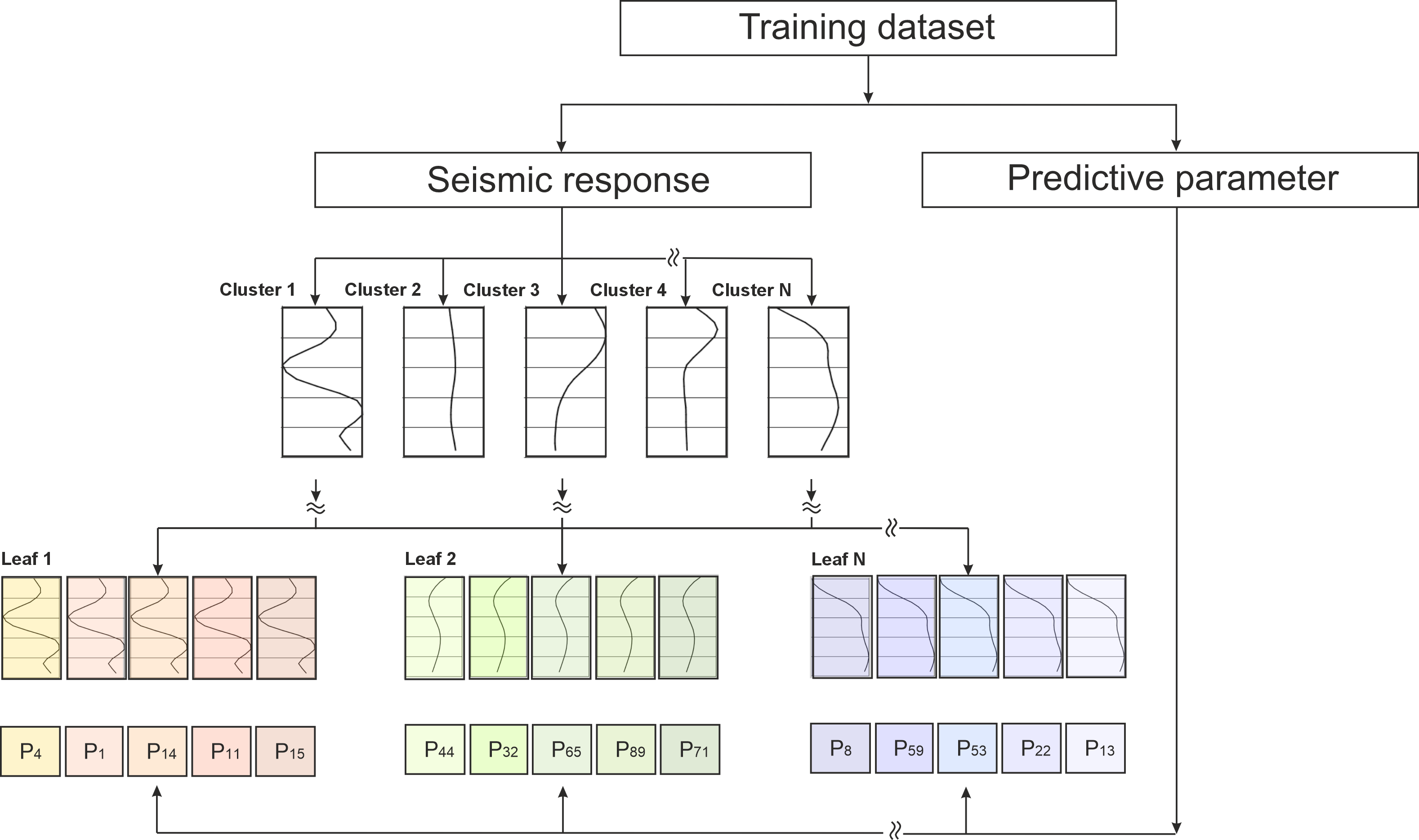}
\caption{Scheme of constructing a clustering decision tree for quantitative interpretation tasks in seismic data analysis.}
\label{fig:clustering_tree}
\end{figure}

A clustering decision tree is constructed based on analyzing numerous seismic responses within the training dataset. At the root node, the tree splits into a predefined number of clusters. Each subsequent node is similarly subdivided until reaching the final "leaf" nodes of the tree (Figure \ref{fig:clustering_tree}). Users determine the tree's depth and branching factor based on data characteristics and prediction goals. Binary splitting ensures simplicity and search speed, whereas multi-level splitting into more clusters allows more accurate representation of complex data relationships. The tree depth can be automatically defined by specifying a minimum object count in the leaves, optimizing network structure.

Following the decision tree construction, a detailed analysis of "seismic response—predictive parameter" pairs occurs within tree leaves. If seismic responses exhibit high similarity, predictive parameters within a leaf should also closely match. Employing the k-NN method enhances predictive accuracy by averaging predictive parameters with similarity coefficients and spatial proximity of neighbors. For discrete parameters, such as reservoir type or lithological facies, predictions rely on the most frequent value among nearest neighbors.

Low similarity in predictive parameters within a single leaf indicates weak linkage between seismic responses and predictive parameters, signaling the need to revisit the chosen model or training parameters.

\begin{figure}[ht]
\centering
\includegraphics[width=\textwidth]{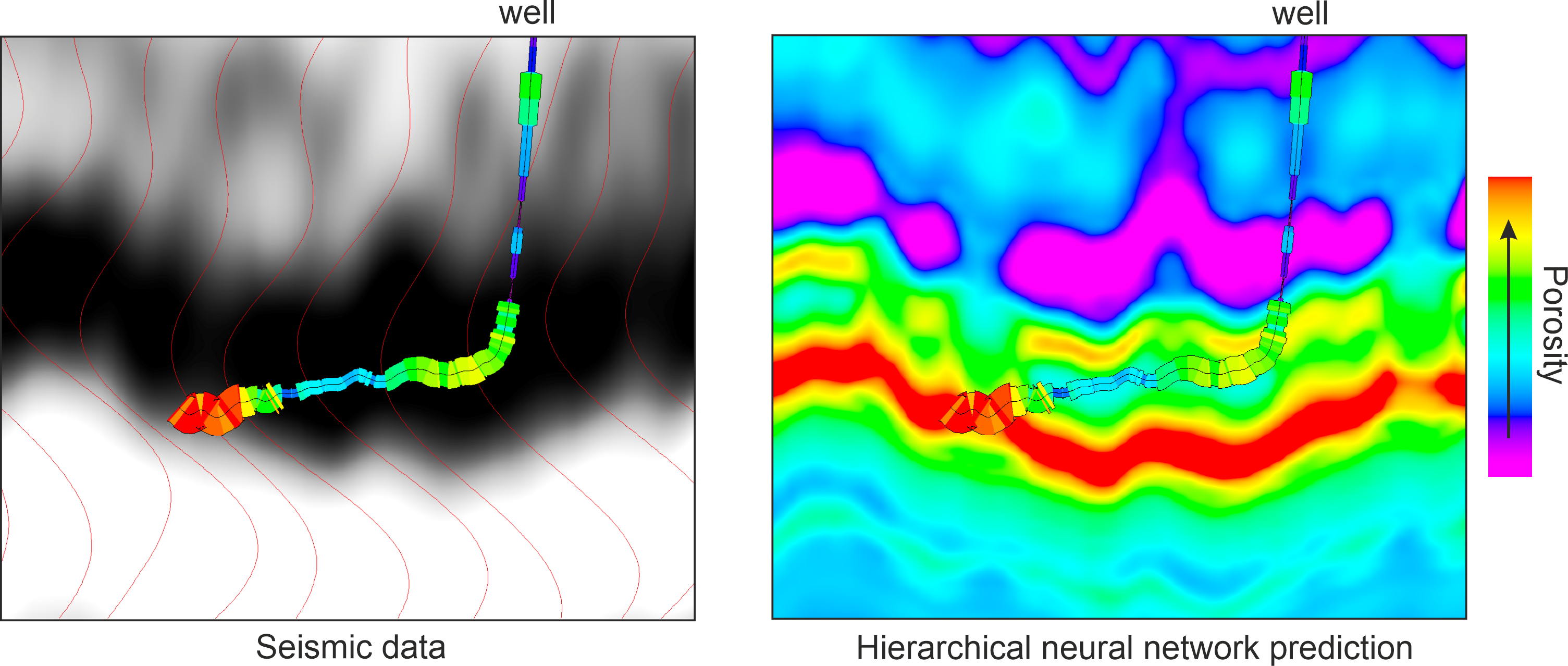}
\caption{Example of porosity prediction using the proposed hierarchical neural network method based on seismic data.}
\label{fig:porosity_prediction}
\end{figure}

During predictions on the basis of new data, original seismic responses traverse the decision tree from the root node to the corresponding leaf, where the predictive parameter is determined (Figure \ref{fig:porosity_prediction}). If a new response significantly deviates from known ones, it is placed into a separate unlabeled cluster. Subsequent responses similar to this new cluster are grouped accordingly, automatically fine-tuning the network and adapting the model to new, previously unknown data types.

Thus, hierarchical neural networks effectively solve analysis and prediction tasks in the oil and gas industry, ensuring high accuracy, flexibility, and continuous learning capabilities.

We also believe that the proposed technology can be successfully applied in various fields where erroneous ("hallucinatory") results from traditional neural network approaches are critically unacceptable:

\begin{itemize}
\item \textbf{Medical diagnostics} based on the analysis of large datasets. In this field, avoiding incorrect diagnoses due to hallucination effects is critically important. The proposed technology can identify multiple sets of medical data (e.g., tests and studies) closely matching specific diagnoses. Comparative analysis of the similarity between new test data and previously established cases (nearest neighbors) allows more accurate assessment of diagnosis probabilities and reduces diagnostic errors \cite{esteva2017dermatologist}.

\item \textbf{Analysis of scientific data and research results} from extensive databases. The proposed approach effectively identifies the uniqueness and novelty of research findings, avoiding duplication and enhancing the quality of scientific conclusions \cite{jordan2015machine}.

\item \textbf{Creation of machine translation} systems based on large parallel text databases in different languages. Effective comparison of textual sentences requires preliminary vectorization using methods such as embedding and attention mechanisms. The proposed technology enhances translation accuracy and minimizes the likelihood of errors associated with hallucination effects \cite{vaswani2017attention}.

\item \textbf{Development of specialized LLMs} for generating and predicting contextual sentence continuations. Such systems require text data vectorization through embedding and attention mechanisms. The proposed method significantly reduces hallucination probabilities and ensures simplicity and effectiveness in model fine-tuning without the risk of catastrophic forgetting, which is crucial when continuously receiving new textual data \cite{brown2020language}.

\item \textbf{Management of critical technological processes}, where errors from traditional neural networks could lead to severe consequences, such as in nuclear energy, chemical industries, and other similar sectors. In such contexts, it is particularly important to ensure high reliability in predicting technological process states, minimizing the risk of hallucination effects.
\end{itemize}

The proposed technology can also be adapted for solving a wide range of other tasks, where traditional neural network applications are limited by hallucination issues and continuous learning challenges.

\section{Discussion}

It is evident that analogies form the foundation of our cognitive processes. When encountering new objects or phenomena, we inevitably attempt to find analogies among previously known images stored in our memory. If sufficient similarity between a new object or phenomenon and known entities is found, we reinforce and validate our existing knowledge. Conversely, when analogies are absent or only distantly related connections are identified, we perceive these objects or phenomena as novel and unique. These instances become the foundation for the creation of entirely new knowledge and concepts.

In the context of understanding the world, human cognition tends to group and classify information. We systematically categorize similar objects or phenomena, forming generalized categories that significantly simplify and accelerate subsequent learning processes. To determine if a presented phenomenon or object is truly novel, we compare it with previously formed hierarchical knowledge structures possessing varying generalization levels. These structures typically exhibit an expanding form, where from the most general concepts ("root nodes"), the hierarchy gradually refines into more specific and concrete elements stored in the "leaves" of such structures. This architecture allows for the rapid and effective comparison and identification of newly incoming data.

The application of mathematical methods based on hierarchical neural network approaches and the nearest neighbor algorithm appears promising for deeper investigation and modeling of human cognitive processes. Specifically, such neural network models allow structuring data in a manner analogous to human brain function, evaluating the similarity of new data against existing information using distance or similarity measures. By employing hierarchical neural networks, we can approach an understanding of how the human brain categorizes knowledge, stores information, and retrieves necessary analogies for comprehending the surrounding world. Neural network applications may also shed light on mechanisms underlying the emergence of new concepts and creative thinking, grounded in discovering unexpected analogies and connections among previously unrelated phenomena and objects.

Further advancement of mathematical and computational models of cognition through hierarchical neural networks could enhance technological approaches to data processing and bring us closer to unraveling fundamental questions about the nature of human intelligence and cognition.

\section{Conclusion}

This study presents an alternative approach to artificial intelligence tasks using hierarchical neural networks combined with the nearest neighbor algorithm. Although the nearest neighbor algorithm is simpler and potentially less innovative compared to classical neural networks, its application can significantly reduce or entirely eliminate the hallucination effects inherent in modern deep neural networks. This characteristic is crucial in fields where high reliability of results is essential.

The key distinction between neural networks and the nearest neighbor algorithm lies in the fact that neural networks aim to generalize and approximate training data through extensive interconnected parameters, requiring substantial computational and time resources. In contrast, the nearest neighbor method evaluates each pair of training data independently, enabling stable and reliable performance even with conflicting datasets. This approach ensures transparency in decision-making and straightforward interpretation of results, as outcomes are directly associated with specific examples from the training set.

The proposed method demonstrates high computational efficiency, significantly surpassing naive implementations of the k-NN algorithm and matching classical neural network models in prediction and classification tasks. The algorithm for constructing a search tree based on Kohonen self-organizing maps structures data effectively, substantially reducing the search time for nearest analogs.

The presented approach is particularly promising in text data processing, where embedding technologies and attention mechanisms are essential. Hierarchical networks ensure linear computational cost scaling relative to training data size and network complexity, distinguishing this method advantageously from traditional neural networks that require substantial computing power for training and inference.

Thus, the proposed approach not only addresses fundamental limitations of traditional neural network technologies but also opens new possibilities for creating robust, interpretable, and reliable artificial intelligence systems. An additional benefit is the proximity of the proposed algorithm to human cognitive principles, which similarly rely on analogies and hierarchical knowledge structuring. This positions the method for successful application not only in practical tasks but also in fundamental research into human intelligence and cognition mechanisms.

\section*{Acknowledgments}

We are grateful to our colleagues from IPLab LLC for their valuable collaboration and support throughout this research.

\section*{Data Availability Statement}

The MNIST dataset used in this study is publicly available. The OpenSubtitles dataset is available through the OPUS project. Code and additional experimental data will be made available upon reasonable request to the corresponding author.

\section*{Conflict of Interest Statement}

The authors declare that they have no competing interests.

\end{document}